%% file: arxiv.tex
\documentclass[runningheads]{llncs}

\input{math_cmds}

\usepackage{todonotes}
\usepackage{amsmath}
\usepackage{multicol,multirow}
\usepackage{subcaption}
\usepackage{graphicx}
\usepackage{highlight}
\renewcommand{\opacity}{50}
\usepackage{tikz}
\usepackage{collcell}

\usepackage{mathtools}

\usepackage{booktabs,siunitx,array,threeparttable,caption}
\usepackage{dcolumn}
\usepackage{xspace}
\usepackage{makecell}
\usepackage{colortbl}
\usepackage{tabularx}
\usepackage{amssymb}
\usepackage[pagebackref,breaklinks,colorlinks]{hyperref}

\usepackage{cleveref}
\usepackage{enumitem} 
\usepackage[version=4]{mhchem}

\usepackage{collcell}

\DeclareRobustCommand\onedot{\futurelet\@let@token\@onedot}
\def\onedot{\ifx\@let@token.\else\null\fi\xspace}

\def\eg{\emph{e.g.}\onedot} 
\def\ie{\emph{i.e.}\onedot} 
 
 \def\vs{\emph{vs}\onedot}
\def\wrt{w.r.t\onedot}

\crefname{figure}{Fig.}{Figs.}
\Crefname{figure}{Figure}{Figures}
\crefname{table}{Table}{Tables}
\Crefname{table}{Table}{Tables}

\input{commands}

\definecolor{myblue}{rgb}{0., 0., 0.9}
\definecolor{myred}{rgb}{0.9803921568627451, 0.4470588235294118, 0.40784313725490196}
\definecolor{occR}{rgb}{1,0,0}
\definecolor{occG}{rgb}{0,255,0}
\definecolor{recon}{HTML}{FF6666}
\definecolor{GT}{HTML}{009900}

\newcommand{\occRed}[1]{\textcolor{occR}{#1}}
\newcommand{\recon}[1]{\textcolor{recon}{#1}}
\newcommand{\gtINR}[1]{\textcolor{GT}{#1}}

\def\methodname{\emph{TrIND}\onedot}

\begin{document}
\title{\methodname: Representing Anatomical \underline{Tr}ees by Denoising \underline{D}iffusion of \underline{I}mplicit \underline{N}eural Fields}

\titlerunning{\methodname}
\author{Ashish Sinha \and 
Ghassan Hamarneh}
\authorrunning{Sinha et. al.}

\institute{Simon Fraser University \\
\email{\{ashish\_sinha,hamarneh\}@sfu.ca}\\
}
\maketitle

\begin{abstract}
    Anatomical trees play a central role in clinical diagnosis and treatment planning. 
    However, accurately representing anatomical trees is challenging due to their varying and complex topology and geometry.
    Traditional methods for representing tree structures, captured using medical imaging, while invaluable for visualizing vascular and bronchial networks, exhibit drawbacks in terms of limited resolution, flexibility, and efficiency.
    Recently, implicit neural representations (INRs) have emerged as a powerful tool for representing shapes accurately and efficiently.
    We propose a novel approach, \methodname, for representing anatomical trees using INR, while also capturing the distribution of a set of trees via denoising diffusion in the space of INRs. 
    We accurately capture the intricate geometries and topologies of anatomical trees at any desired resolution.
    Through extensive qualitative and quantitative evaluation, we demonstrate high-fidelity tree reconstruction with arbitrary resolution yet compact storage, and versatility across anatomical sites and tree complexities.
    The code is available at: \texttt{\url{https://github.com/sinashish/TreeDiffusion}}.
\end{abstract}


\section{Introduction}
\label{sec:intro}

\begin{figure}[h] 
\centering 
\begin{subfigure}{\textwidth} 
\centering 
    \begin{subfigure}{0.49\linewidth}
        \centering
        \includegraphics[width=0.9\linewidth, keepaspectratio]{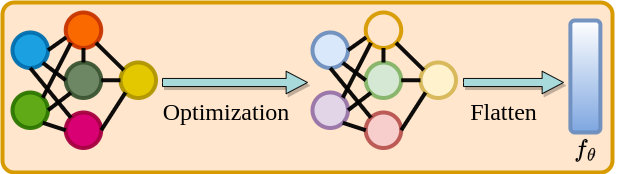} 
        \caption{Per-instance optimization}
        \label{fig:overfitting}
    \end{subfigure}
    \begin{subfigure}{0.49\linewidth}
        \centering
        \includegraphics[width=0.9\linewidth, keepaspectratio]{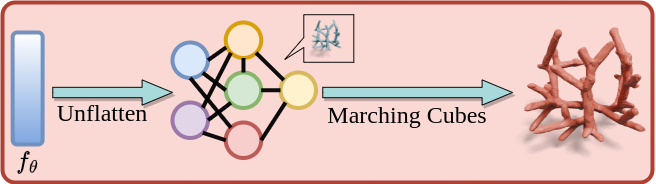} 
        \caption{Isosurface extraction from INR}
        \label{fig:mesh_extraction}
    \end{subfigure}
    
\end{subfigure} 
\begin{subfigure}{\textwidth} 
    \centering 
    \begin{subfigure}{0.49\linewidth}
        \centering
        \includegraphics[width=\linewidth, height=0.3\linewidth,keepaspectratio]{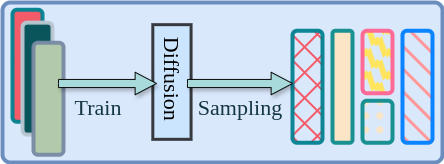} 
        \caption{Learn tree-space via denoising diffusion}
        \label{fig:ddim_overview}
    \end{subfigure}
    \begin{subfigure}{0.49\linewidth}
        \centering
        \includegraphics[width=0.9\linewidth, keepaspectratio]{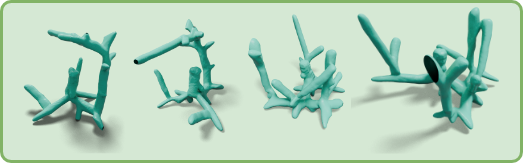}
        \caption{Generate novel tree structures}
        \label{fig:ddim_teaser}
    \end{subfigure}
\end{subfigure} 
        
    \caption{\textbf{\methodname overview}: (a) An INR is optimized for each sample in the training dataset, and then flattened to a 1D vector. (b) During inference, the INR is recovered from the flattened vector, followed by MC to extract the underlying signal. (c) The diffusion transformer takes the flattened vectors as input to model the diffusion process. After training, novel INRs can be sampled and used for downstream tasks via (b).
    (d) Novel tree structures visualized as mesh.
    }
\end{figure}

\noindent \paragraph{\textbf{Motivation for Studying Anatomical Trees.}}
Accurate extraction and analysis of anatomical trees, including vasculature and airways, is crucial for predicting fatal diseases like ischemic heart disease~\cite{who} and aiding clinical tasks like surgical planning~\cite{sobocinski2013benefits}, computational fluid dynamics (CFD)~\cite{tran2021patient}, and disease prognosis~\cite{alblas2022going}.
However, the application of extracted trees for such purposes may be hindered by their explicit representation as meshes or volumetric grids.
\noindent \paragraph{\textbf{Previous Works on Tree Representation.}} 
Prior works have utilized discrete and explicit representations, such as medial axis~\cite{pizaine2011implicit}, minimal paths~\cite{li2007vessels}, grammar~\cite{lindenmayer1968mathematical}, voxel grids~\cite{zhao2019tree}, 
and meshes~\cite{yang2020intra}, to represent anatomical trees.
However, all these explicit representations have their own peculiar shortcomings, requiring complex ad-hoc machinery~\cite{hu2022subdivision,qi2016volumetric} to obtain a smooth surface essential for vascular modeling~\cite{alblas2022going}, and/or large memory footprint~\cite{maturana2015voxnet}, up to O$(n^3)$ for voxels, to preserve fine details.
Although rule-based representations, such as L-system, offer some degree of control over the produced structures~\cite{prusinkiewicz1994synthetic}, the complexity of rules and parameters and the computational cost of generating and rendering the shapes limit their use in current medical applications.
Therefore, there is a need for a continuous shape representation that is independent of spatial resolution, is memory efficient, and can easily be integrated into deep learning pipelines.

\noindent \paragraph{\textbf{Implicit Neural Representation (INR).}}
Recently, INRs (or neural fields, implicit fields) have been proposed to address the shortcomings of explicit representations, which use multi-layer perceptrons (MLP) to fit a continuous function that \emph{implicitly} represents a signal of interest~\cite{mescheder2019occupancy,park2019deepsdf,sitzmann2020implicit} as level sets.
This approach not only learns a representation of arbitrary shapes with a small number of parameters while still achieving a high surface accuracy, but also supports reconstructions at any arbitrary resolution.
To this end, we propose to use INRs for representing anatomical trees and demonstrate their efficiency and effectiveness in modeling simple and complex, vasculature and airways, in 2D and 3D, on both synthetic and real data.

\noindent \paragraph{\textbf{Generative Modeling of Tree Distributions.}}
While implicit representations and level sets have been widely used for segmentation~\cite{van2002level,lorigo2001curves} and 3D modeling~\cite{hong2020high,kretschmer2013interactive} of vessel structures, there are limited works that address the modeling of the ``shape space" of tree structures.
Previous works that attempt to analyze and synthesize tree structures are either \emph{model-based} or \emph{data-driven}.
\emph{Model-based} approaches~\cite{galarreta2013three,hamarneh2010vascusynth,talou2021adaptive,zamir2001arterial} record the tree topology along with branch attributes~\cite{feragen2012toward,zhao2017leveraging}, \eg, length and angle. However, for trees with diverse topologies (\eg, different number of branches), a fixed-size representation no longer suffices, and altering the model size across trees impairs statistical analysis. 
\emph{Data-driven} methods use generative modeling to represent the distribution of training data and sample the distribution to synthesize novel instances.
GANs and VAEs were used to model the distribution of vessels~\cite{feldman2023vesselvae,wolterink2018blood} but did not leverage the descriptive power of INRs as~\cite{alblas2022going,park2019deepsdf} do.
Diffusion models~\cite{ho2020denoising} were employed in~\cite{chou2023diffusion} to learn the distribution on latent VAE codes of shapes but required large training sets for faithful encoding.
For a more faithful representation, INRs were used in~\cite{erkocc2023hyperdiffusion}, though neither~\cite{chou2023diffusion,erkocc2023hyperdiffusion} modeled tubular structures or trees.

\noindent \paragraph{\textbf{Our contributions.}} 
To address the shortcomings of previous works, we are the first to make the following contributions:
(1) We train INRs to achieve a faithful representation of complex anatomical trees and demonstrate their usage in segmenting trees from medical images;
(2) we employ diffusion models on the space of INR-represented trees for learning tree distributions and generating plausible novel trees with complex topology;
(3) we showcase the versatility of \methodname in representing trees of varying dimensionality, complexity, and anatomy; and
(4) We quantitatively assess our method's compactness representation and reconstruction accuracy, at arbitrarily high resolution.

\begin{figure}[h]
    \centering
    \begin{subfigure}{\textwidth} 
        \centering 
        \includegraphics[width=0.9\linewidth]{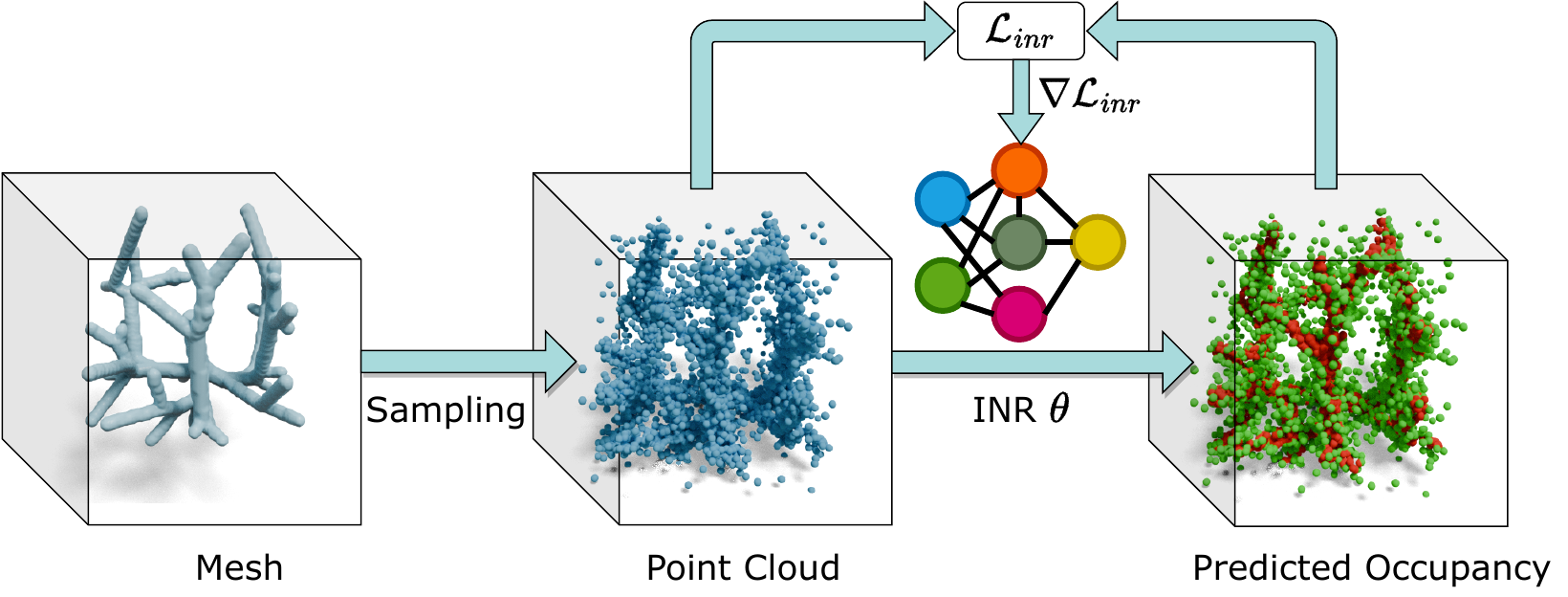} 
        \label{fig:optimization}
    \end{subfigure} 
    \makebox[\textwidth]{}{}
    \begin{subfigure}{\textwidth} 
        \centering  
        \includegraphics[width=0.9\linewidth]{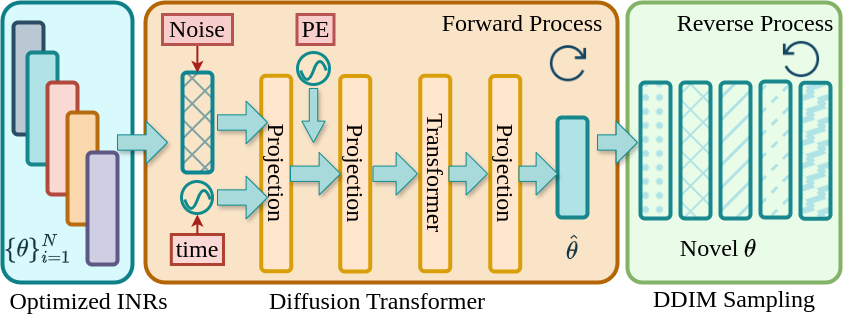}
        
        \label{fig:ddim}
    \end{subfigure} 
    \caption{\textbf{Pipeline}: Top: Given a 3D mesh, sampled points and GT occupancies (\occRed{inside}/\gtINR{outside}), an INR $(\theta)$ is optimized to fit to the shape. Bottom: Optimized INRs are flattened to a 1D vector, fed to the transformer-based diffusion model $\mathcal{D}(\phi)$ and optimized to predict noise. The novel INRs can then be sampled from the trained $\mathcal{D}(\phi)$ in the reverse process.}
    \label{fig:pipeline}
\end{figure}

\section{Methodology}
\noindent \paragraph{\textbf{Overview.}} 
We propose \methodname (pronounced as \emph{Trendy}), a diffusion-based generative approach comprising two steps.
First, we optimize an INR realized as an MLP by overfitting its parameters to each tree in a dataset so that each INR faithfully represents each tree as a neural occupancy field~(\cref{fig:overfitting}).
Second, the optimized INRs are flattened into 1D vectors and used to train a denoising diffusion model (DDM) that captures the statistical distribution of the trees~(\cref{fig:ddim_overview}). 
The DDM is used to synthesize new INRs, via a reverse diffusion process~(\cref{fig:mesh_extraction}), 
which represents the occupancy field (or a mesh, if marching cubes~\cite{lorensen1998marching} is applied) of the tree~(\cref{fig:ddim_teaser}).

\noindent \paragraph{\textbf{Stage 1: Learning per-sample INR.}}
\label{sec:stage1}

We represent each tree, in a dataset $\{S_{i}\}_{i=1}^N$ of $N$ trees, as a neural occupancy field, or INR,  $f(x;\theta) : \mathbb{R}^3 \to [0,1]$, modelled as an MLP and parameterized by $\theta \in \mathbb{R}^D$.
Note that, unlike prior works~\cite{park2019deepsdf,zhang20233dshape2vecset} that involve parameter sharing across the entire dataset, we optimize an INR separately for each tree sample.

Given $o_i(x)$, the ground truth occupancy of $x$ for $S_{i}$, we optimize $\theta_i$ as:
\begin{equation}
    \argmin_{\theta_i} \sum_{x_j \in \mathbb{R}^3} \| f_{\theta_i}(x_j) - o_i(x_j) \|_2.
    \label{eq:mse_loss}
\end{equation}

\noindent Similar to~\cite{mescheder2019occupancy,niemeyer2020differentiable}, our MLP architecture is a fully-connected network with $L$ layers, of hidden size $D$,
and ReLU activation functions.
To learn high-fidelity neural representations of anatomical trees, we follow~\cite{mescheder2019occupancy} to first instantiate all training instances $S_{i}$ in Lipschitz domain $\Omega \in [-1, 1]$ and then adaptively sample points and ground-truth occupancies both inside and outside the 3D surface of $S_i$, to supervise the optimization of each MLP until convergence.
\Cref{fig:pipeline} illustrates this optimization process.

\noindent \paragraph{\textbf{Stage 2: Diffusion on INRs.}}
\label{sec:stage2}
We train a diffusion model on the space of INRs, \ie, the weights and biases $\{ \theta_i \}_{i=1}^N$. 
Our transformer-based~\cite{vaswani2017attention} diffusion model $\mathcal{D}(\phi)$ is inspired by~\cite{erkocc2023hyperdiffusion,peebles2022learning}, and 
takes the flattened 1D vectors of $\theta_i$ as input.
However, before passing the input to $\mathcal{D}(\phi)$, each $\theta_i$ undergoes a \emph{layer-by-layer} decomposition into $k$ tokens by MLPs~\cite{peebles2022learning}.
This is essential due to the potential variation in the dimensionality of different layers in $\theta_i$ and ensures correspondence of any element $j$ across all trees $\theta_{i}, \forall i,j$.
During forward diffusion, we apply noise $\eta_t$ at timestep $t$ to each vector $\theta_i$ to obtain a noisy vector $\theta_i^{\star}$.

Following~\cite{erkocc2023hyperdiffusion,peebles2022learning}, the sinusoidal embedding of $t$ along with $\theta_i^{\star}$ is fed to a linear projection layer, whose output is concatenated with a learnable positional encoding vector to obtain $\mathcal{G}$. 
The resultant $\mathcal{G}$ is then passed to the transformer, which outputs denoised INRs $\hat{\theta_i}$.
Parameters $\phi$ of $\mathcal{D}$ are optimized using:
\begin{equation}
    \label{eq:diff_phi_mse}
    \argmin_\phi \frac{1}{N} \sum_{i=1}^N \| \hat{\theta_i} - \theta_i \|_2.
\end{equation}
After training, we employ DDIM~\cite{song2020denoising} sampling, \ie, a reverse diffusion process, to sample new INRs by feeding-in noise as $\hat{\theta}_i$ to $\mathcal{D}(\phi)$, and gradually denoising it.
\Cref{fig:ddim} shows an illustration of the overall diffusion process.

\section{Datasets, Results, and Implementation Details}
\label{sec:results}

\paragraph{\textbf{Datasets.}}
We use the following datasets to evaluate our approach: 
(1) \underline{VascuSynth}: 120 synthetic 3D vascular trees, ranging from 1 branch to complex trees with multiple branches and bifurcations~\cite{hamarneh2010vascusynth}; 
(2) \underline{IntRA}: 103 3D meshes of intracranial vasculature extracted from MRA~\cite{yang2020intra};
(3) \underline{BraTS}: 40 3D MRI brain scans~\cite{menze2014multimodal};
(4) \underline{HaN-Seg}: 20 segmentations of Head\&Neck CTA scans~\cite{podobnik2023han}; 
(5) \underline{DRIVE}: 20 2D retinal fundus color images with vessel segmentation masks~\cite{staal2004drive}; 
(6) \underline{EXACT}: One CT scan of a bronchial tree~\cite{lo2012exact09};
(7) \underline{WBMRA}: One whole body MRA; and
(8) \underline{CoW}: One circle of Willis mesh. 
Our evaluation covers the following aspects.

\begin{figure}[t]
    \centering
    \begin{subfigure}{\textwidth}
        \centering
        \includegraphics[width=\linewidth, keepaspectratio]{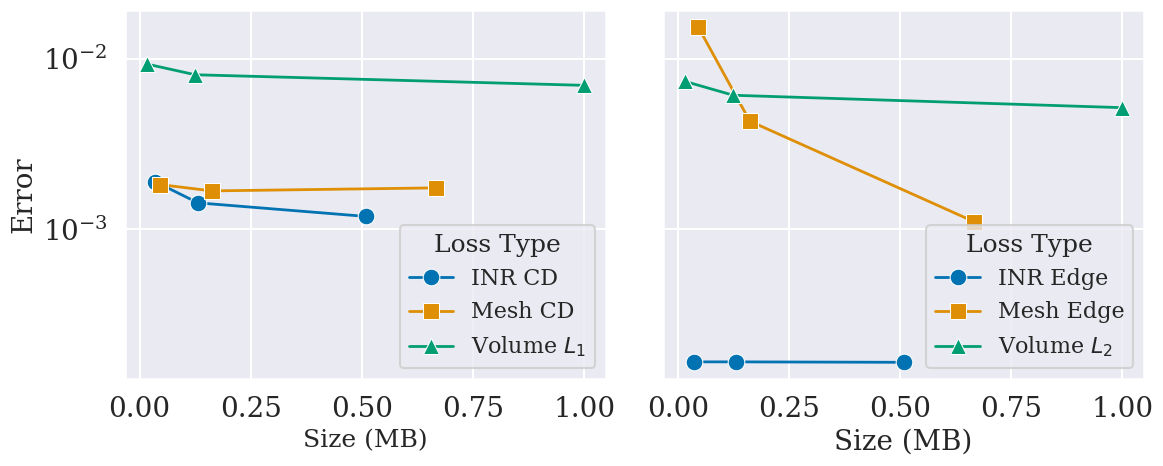}
        \caption{}
    \end{subfigure}
    \begin{subfigure}{\textwidth}
        \centering
        \includegraphics[width=0.24\linewidth]{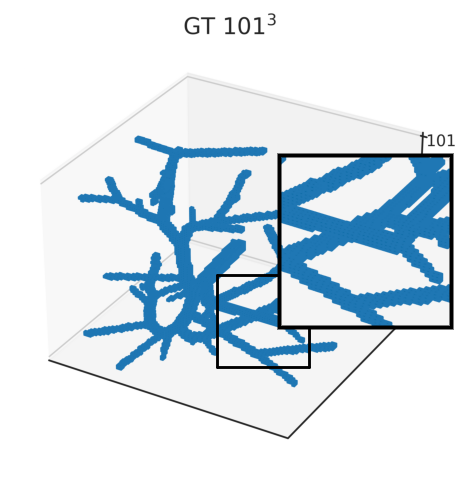}
        \includegraphics[width=0.24\linewidth]{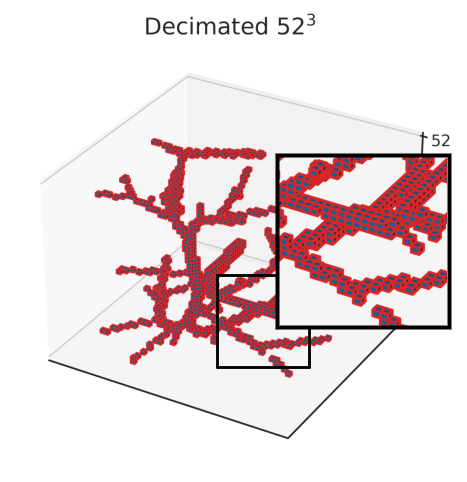}
        \includegraphics[width=0.24\linewidth]{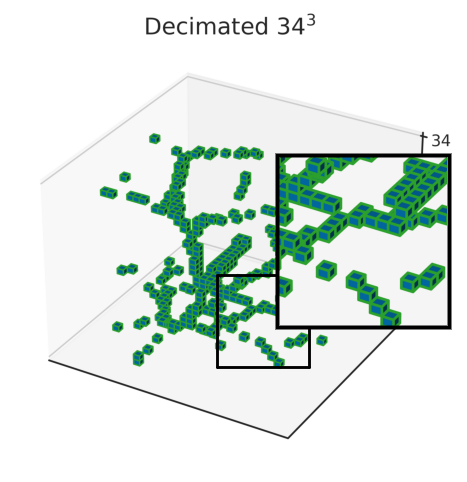}
        \includegraphics[width=0.24\linewidth]{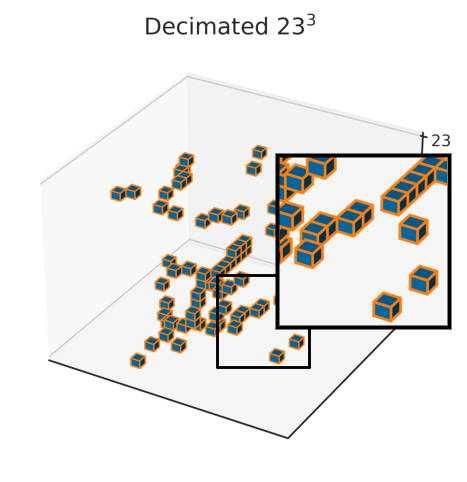}
        \caption{}
    \end{subfigure}
    \caption{\textbf{Compression \vs Reconstruction Accuracy}: (a) We decimate/downscale the mesh/volume to occupy $\approx$ same memory space as INR, and report the reconstruction error using Chamfer distance (CD) and edge length loss (Edge) for meshes and INRs, and $L_1$ and $L_2$ for volumes. Note the higher error for meshes and volumes \wrt INRs for the same storage. (b) Illustration of GT and downsampled volumes. Notice the disconnected components.}
    \label{fig:size_vs_error}
\end{figure}

\noindent \paragraph{\textbf{Fidelity and Compactness.}}

Following~\cite{mescheder2019occupancy}, we use Chamfer distance (CD) to evaluate the fidelity of INRs in representing anatomical trees
and summarize it in~\cref{tab:modality_inr,tab:vasc_mlp_dim}.
\Cref{fig:size_vs_error} shows that we can achieve higher reconstruction accuracy with a smaller memory footprint.
For example, as seen in~\cref{tab:modality_inr} we need 68 MB for volumes and 12 MB for meshes compared to INRs that only occupy 0.75 MB and 0.63 MB, respectively, offering $\approx90\times$ and $\approx19\times$ compression, respectively, with minimal loss of reconstruction accuracy.
Similarly, \cref{tab:vasc_mlp_dim} shows that $10k$ parameters of INR are enough to encode the geometry of a $128^3$ volume of IntRA, resulting in $\approx220\times$ compression.

\begin{table}[h]
    \centering
    \caption{Quantitative comparison of reconstruction accuracy and compression ratio on the VascuSynth (\textbf{V})~\cite{hamarneh2010vascusynth} and IntRA (\textbf{I})~\cite{yang2020intra} for different configurations of our MLP architecture. CD denotes chamfer distance ($10^{-3}$) between the reconstructed mesh and the ground truth, whereas compression denotes the ratio between size of INR and size of the original input modality, \ie, raw volume. $\uparrow$ denotes higher is better and $\downarrow$ lower is better.}
    \label{tab:vasc_mlp_dim}
    \resizebox{\textwidth}{!}{
    \large
    \begin{tabular}{lcccccccccccccccccc}
        \toprule
        & \multicolumn{4}{c}{$L=1$} & & & \multicolumn{4}{c}{$L=3$} & & & \multicolumn{4}{c}{$L=5$} \\
        \cmidrule[0.5pt](rl){2-7}
        \cmidrule[0.5pt](rl){8-13}
        \cmidrule[0.5pt](rl){14-19}
        & & & \multicolumn{2}{c}{\makecell{Compression ($\uparrow$)}} & \multicolumn{2}{c}{CD ($\downarrow$)} & & &   \multicolumn{2}{c}{\makecell{Compression ($\uparrow$)}} & \multicolumn{2}{c}{CD ($\downarrow$)}  & & &  \multicolumn{2}{c}{\makecell{Compression ($\uparrow$)}} & \multicolumn{2}{c}{CD ($\downarrow$)}  \\
        \cmidrule[0.45pt](rl){4-5}
        \cmidrule[0.45pt](rl){6-7}
        \cmidrule[0.45pt](rl){10-11}
        \cmidrule[0.45pt](rl){12-13}
        \cmidrule[0.45pt](rl){16-17}
        \cmidrule[0.45pt](rl){18-19}
       $D$ & $\#$Params (M) & $\text{Size}_{{inr}}$  (MB) & \makecell{\textbf{V}} & \makecell{\textbf{I}} & \makecell{\textbf{V}} & \makecell{\textbf{I}} &  $\#$Params (M) & $\text{Size}_{{inr}}$   (MB) & \makecell{\textbf{V}} & \makecell{\textbf{I}} &  \makecell{\textbf{V}} & \makecell{\textbf{I}} &  $\#$Params (M) & $\text{Size}_{{inr}}$   (MB) & \makecell{\textbf{V}} & \makecell{\textbf{I}} & \makecell{\textbf{V}} & \makecell{\textbf{I}} \\
        \midrule
        64 & 0.01 & {\cellcolor[HTML]{F7FCF0}} \color[HTML]{000000} 0.03 & {\cellcolor[HTML]{7F0000}} \color[HTML]{F1F1F1} 27.05 & {\cellcolor[HTML]{7F0000}} \color[HTML]{F1F1F1} 223.57 & {\cellcolor[HTML]{084081}} \color[HTML]{F1F1F1} 12.95 & {\cellcolor[HTML]{6FC5C8}} \color[HTML]{000000} 29.84 & 0.03 & {\cellcolor[HTML]{F7FCF0}} \color[HTML]{000000} 0.10 & {\cellcolor[HTML]{7F0000}} \color[HTML]{F1F1F1} 9.47 & {\cellcolor[HTML]{7F0000}} \color[HTML]{F1F1F1} 78.25 & {\cellcolor[HTML]{084081}} \color[HTML]{F1F1F1} 8.65 & {\cellcolor[HTML]{4AAFD1}} \color[HTML]{F1F1F1} 15.48 & 0.04 & {\cellcolor[HTML]{F7FCF0}} \color[HTML]{000000} 0.16 & {\cellcolor[HTML]{7F0000}} \color[HTML]{F1F1F1} 5.73 & {\cellcolor[HTML]{7F0000}} \color[HTML]{F1F1F1} 47.37 & {\cellcolor[HTML]{084081}} \color[HTML]{F1F1F1} 8.74 & {\cellcolor[HTML]{084081}} \color[HTML]{F1F1F1} 21.90 \\
        128 & 0.03 & {\cellcolor[HTML]{F5FBEE}} \color[HTML]{000000} 0.13 & {\cellcolor[HTML]{FDD39D}} \color[HTML]{000000} 7.32 & {\cellcolor[HTML]{FDD39D}} \color[HTML]{000000} 60.50 & {\cellcolor[HTML]{BDE5BE}} \color[HTML]{000000} 10.36 & {\cellcolor[HTML]{C5E8C2}} \color[HTML]{000000} 24.21 & 0.10 & {\cellcolor[HTML]{F6FBEF}} \color[HTML]{000000} 0.38 & {\cellcolor[HTML]{FDD29C}} \color[HTML]{000000} 2.49 & {\cellcolor[HTML]{FDD29C}} \color[HTML]{000000} 20.55 & {\cellcolor[HTML]{C2E7C0}} \color[HTML]{000000} 7.09 & {\cellcolor[HTML]{EBF7E5}} \color[HTML]{000000} 8.33 & 0.17 & {\cellcolor[HTML]{F5FBEE}} \color[HTML]{000000} 0.63 & {\cellcolor[HTML]{FDD29C}} \color[HTML]{000000} 1.50 & {\cellcolor[HTML]{FDD29C}} \color[HTML]{000000} 12.37 & {\cellcolor[HTML]{5EBCCE}} \color[HTML]{000000} 7.99 & {\cellcolor[HTML]{51B5D2}} \color[HTML]{F1F1F1} 17.85 \\
        256 & 0.13 & {\cellcolor[HTML]{ECF8E6}} \color[HTML]{000000} 0.51 & {\cellcolor[HTML]{FFF1DD}} \color[HTML]{000000} 1.88 & {\cellcolor[HTML]{FFF1DD}} \color[HTML]{000000} 15.53 & {\cellcolor[HTML]{F7FCF0}} \color[HTML]{000000} 9.24 & {\cellcolor[HTML]{F7FCF0}} \color[HTML]{000000} 18.19 & 0.40 & {\cellcolor[HTML]{ECF8E6}} \color[HTML]{000000} 1.51 & {\cellcolor[HTML]{FFF0DB}} \color[HTML]{000000} 0.61 & {\cellcolor[HTML]{FFF0DB}} \color[HTML]{000000} 5.06 & {\cellcolor[HTML]{F7FCF0}} \color[HTML]{000000} 6.46 & {\cellcolor[HTML]{F7FCF0}} \color[HTML]{000000} 7.48 & 0.66 & {\cellcolor[HTML]{ECF8E6}} \color[HTML]{000000} 2.51 & {\cellcolor[HTML]{FEEFDA}} \color[HTML]{000000} 0.38 & {\cellcolor[HTML]{FFF0DB}} \color[HTML]{000000} 3.12 & {\cellcolor[HTML]{ECF8E6}} \color[HTML]{000000} 7.07 & {\cellcolor[HTML]{F7FCF0}} \color[HTML]{000000} 11.33 \\
        512 & 0.53 & {\cellcolor[HTML]{CCEBC6}} \color[HTML]{000000} 2.01 & {\cellcolor[HTML]{FFF7EC}} \color[HTML]{000000} 0.47 & {\cellcolor[HTML]{FFF7EC}} \color[HTML]{000000} 3.86 & {\cellcolor[HTML]{C6E9C2}} \color[HTML]{000000} 10.24 & {\cellcolor[HTML]{BFE6BF}} \color[HTML]{000000} 24.66 & 1.58 & {\cellcolor[HTML]{CCEBC6}} \color[HTML]{000000} 6.02 & {\cellcolor[HTML]{FFF7EB}} \color[HTML]{000000} 0.08 & {\cellcolor[HTML]{FFF6E9}} \color[HTML]{000000} 1.29 & {\cellcolor[HTML]{95D6BB}} \color[HTML]{000000} 7.40 & {\cellcolor[HTML]{084081}} \color[HTML]{F1F1F1} 19.99 & 2.63 & {\cellcolor[HTML]{CCEBC6}} \color[HTML]{000000} 10.03 & {\cellcolor[HTML]{FFF6E9}} \color[HTML]{000000} 0.09 & {\cellcolor[HTML]{FFF6E9}} \color[HTML]{000000} 0.77 & {\cellcolor[HTML]{F7FCF0}} \color[HTML]{000000} 6.96 & {\cellcolor[HTML]{E3F4DE}} \color[HTML]{000000} 12.50 \\
        1024 & 2.10 &  {\cellcolor[HTML]{084081}} \color[HTML]{F1F1F1} 8.03 & {\cellcolor[HTML]{FFF5E7}} \color[HTML]{000000} 0.04 & {\cellcolor[HTML]{FFF5E7}} \color[HTML]{000000} 0.96 & {\cellcolor[HTML]{61BDCD}} \color[HTML]{000000} 11.37 & {\cellcolor[HTML]{084081}} \color[HTML]{F1F1F1} 40.06 & 6.30 & {\cellcolor[HTML]{084081}} \color[HTML]{F1F1F1} 24.04 & {\cellcolor[HTML]{FFF7EC}} \color[HTML]{000000} 0.04 & {\cellcolor[HTML]{FFF7EC}} \color[HTML]{000000} 0.32 & {\cellcolor[HTML]{99D7BA}} \color[HTML]{000000} 7.37 & {\cellcolor[HTML]{41A5CB}} \color[HTML]{F1F1F1} 15.87 & 10.50 & {\cellcolor[HTML]{084081}} \color[HTML]{F1F1F1} 42.48 & {\cellcolor[HTML]{FFF7EC}} \color[HTML]{000000} 0.02 & {\cellcolor[HTML]{FFF7EC}} \color[HTML]{000000} 0.19 & {\cellcolor[HTML]{A6DCB6}} \color[HTML]{000000} 7.64 & {\cellcolor[HTML]{65C0CB}} \color[HTML]{000000} 17.26 \\
        \bottomrule
    \end{tabular}
    }
\end{table}

\begin{figure}[h] 
    \centering
    \includegraphics[width=\textwidth]{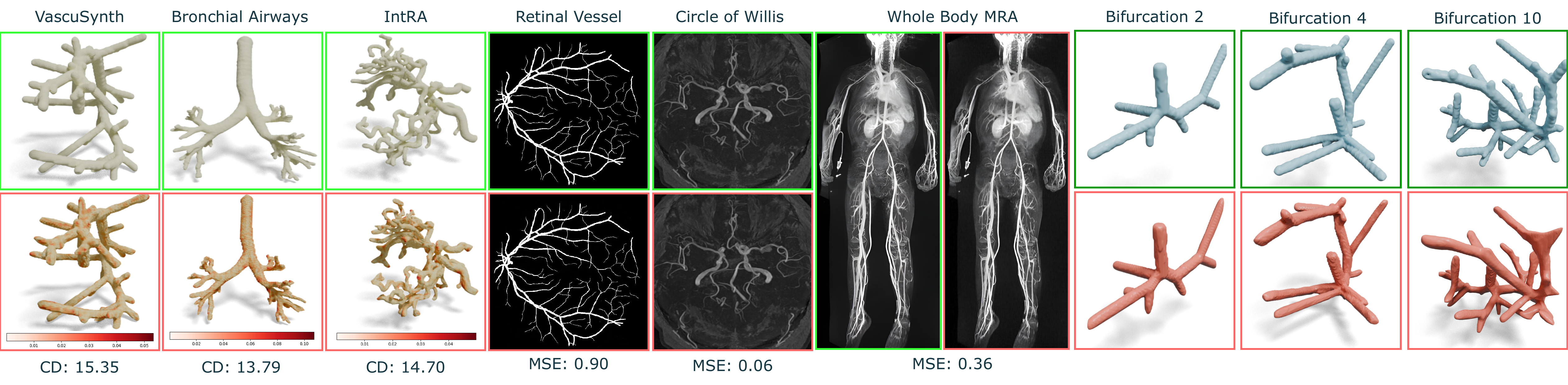}
    \begin{subfigure}{0.68\textwidth}
        \centering
        \caption{Various anatomical sites and imaging modalities}
        \label{fig:inr_modality}
    \end{subfigure}
    \begin{subfigure}{0.28\textwidth}
        \centering
        \caption{Various  complexities}
        \label{fig:inr_bif}
    \end{subfigure}
    
    \caption{\textbf{Versatility}: Visualization of different synthetic and real anatomical trees represented as an INR from various medical imaging modalities and organs. We normalize all shapes to $[-1,1]$ and images to $[0,1]$ to report the reconstruction error using MSE and CD ($\times 10^{-3}$) between \gtINR{ground truth} and the underlying signal extracted from \recon{INR}. 
    } 
\end{figure}

\noindent \paragraph{\textbf{Versatility.}}
We assess the effectiveness of INRs in representing anatomical trees from various medical image modalities.
\Cref{fig:inr_modality,tab:modality_inr} 
qualitatively and quantitatively show the adaptability of INRs in modeling DRIVE vessel masks, 3D meshes, CTA and MRA volumes from VascuSynth, IntRA, EXACT and CoW.
\Cref{fig:inr_bif} depicts how the same MLP architecture can represent trees of varying complexity.

\begin{figure}[!h]
    \centering
    \includegraphics[width=0.95\textwidth]{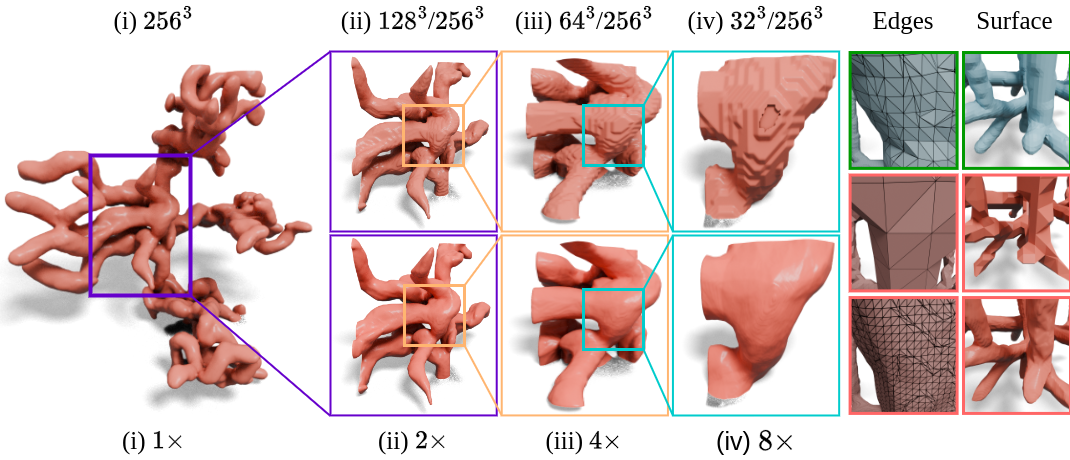}
    \makebox[0.4\textwidth]{}{\scriptsize (a)}
    \makebox[0.35\textwidth]{}{\scriptsize (b)}
    \caption{\textbf{Arbitrary Resolution}: (a) Comparison of 2x, 4x, and 8x zoom on an IntRA sample represented as a volumetric grid (top) and INR (bottom). The resolution for each sub-figure is shown on top as volume$^3$/INR$^3$. Notice the smoothness of the surface even at 8x zoom. (b) Zoomed-in regions of a VascuSynth mesh \recon{reconstructed from INRs} and \gtINR{ground truth} at different mesh resolutions displayed using faces and edges.  }
    \label{fig:vasc_inr_mesh_res}
\end{figure}

\noindent \paragraph{\textbf{Arbitrary Resolution.}} 
Achieving high-quality mesh and volume reconstruction is crucial for the accurate representation of vasculature, \eg, for fluid flow simulation. \Cref{fig:vasc_inr_mesh_res} illustrates how INRs can reconstruct anatomical trees from both IntRA and VascuSynth at various resolutions without the need for adjusting the model size or retraining.

\begin{figure}
    \centering
    \begin{subfigure}{0.49\linewidth}
            \centering            \includegraphics[width=\linewidth,keepaspectratio]{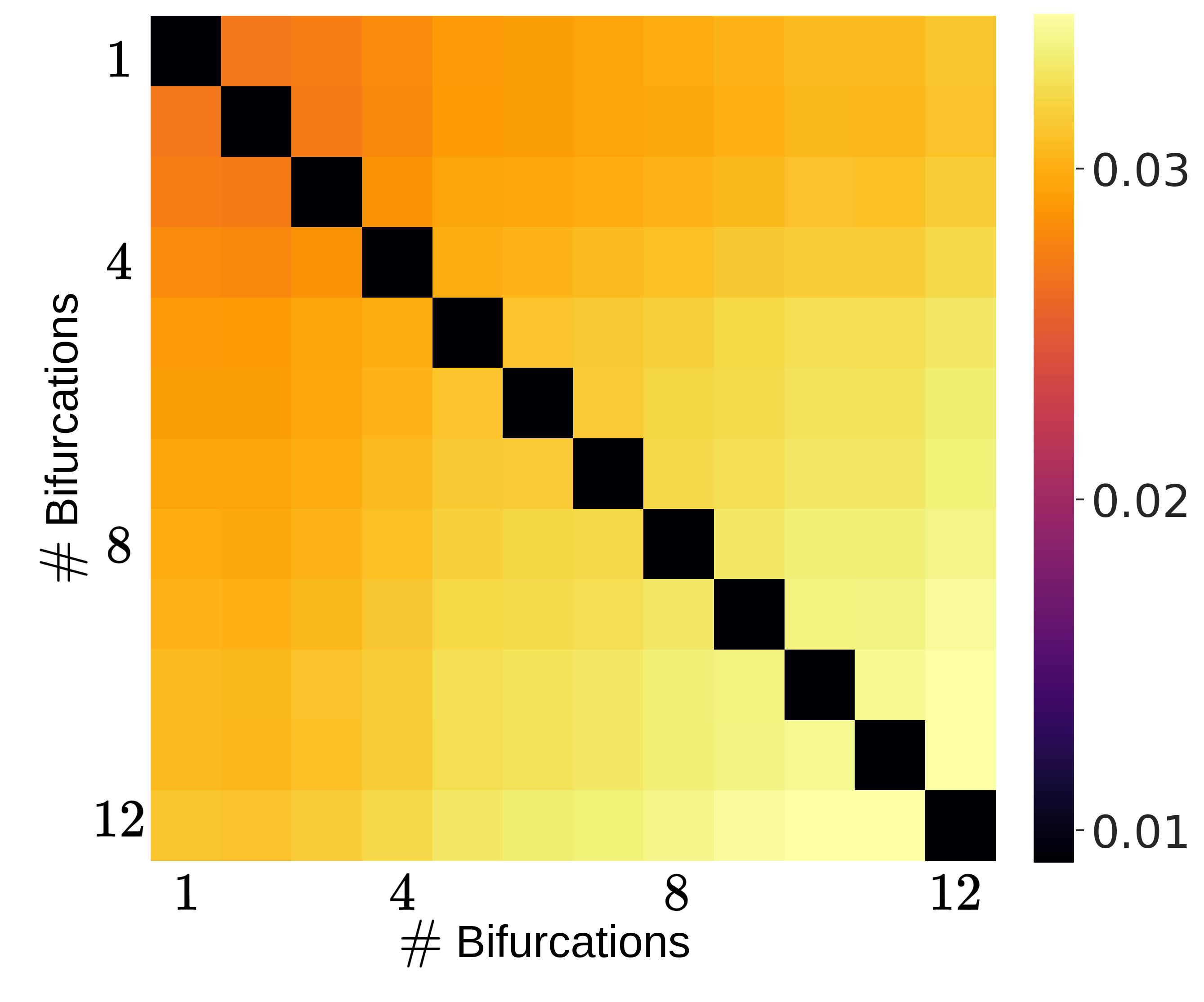}
            \caption{}
            \label{fig:l2_tree}
        \end{subfigure}
    \begin{subfigure}{0.49\linewidth}
            \centering            \includegraphics[width=\linewidth,keepaspectratio,trim={0 0.5cm 0 0.5cm}, clip]{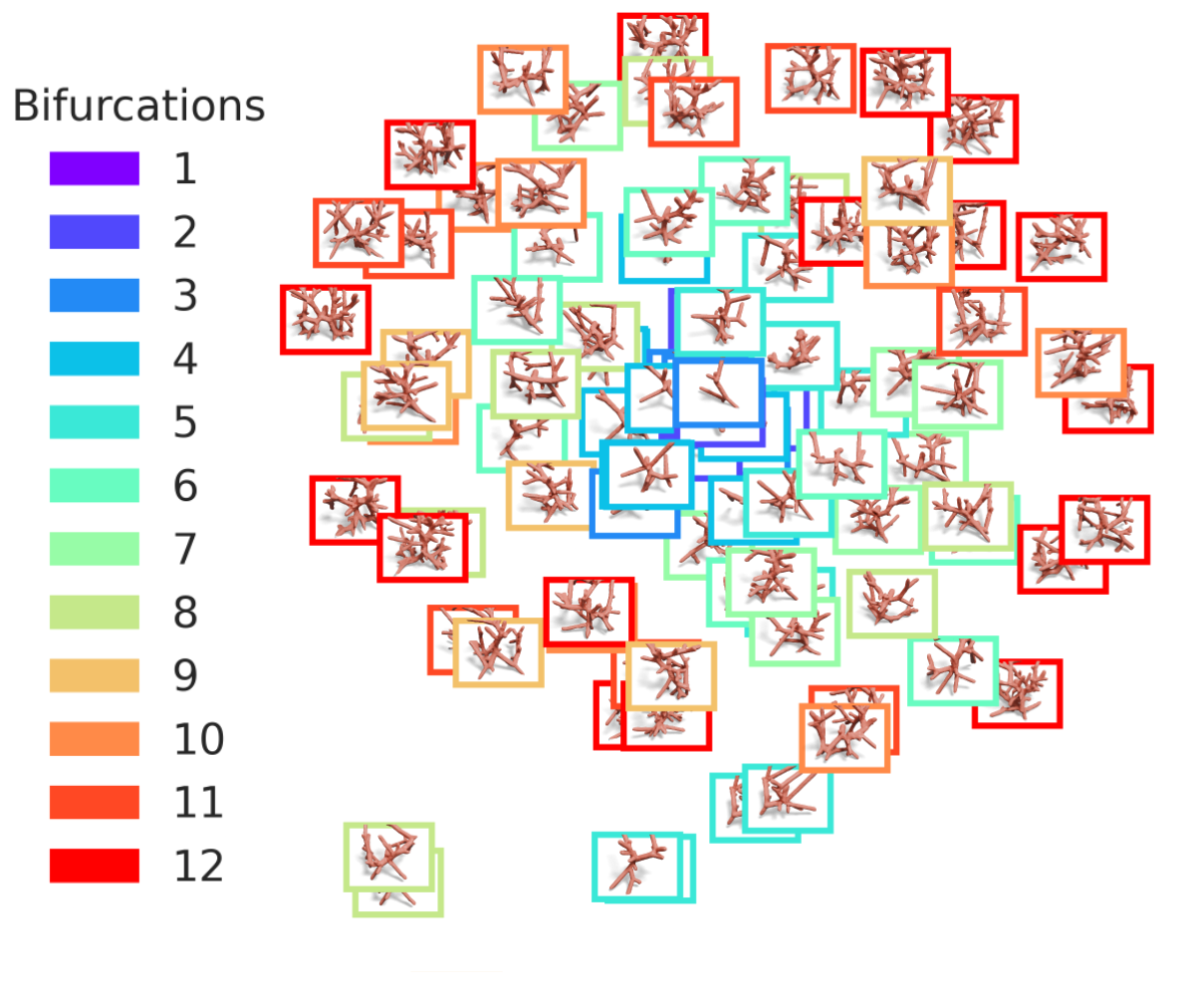}
            \caption{}
            \label{fig:tsne_few}
        \end{subfigure}
        \begin{subfigure}{0.49\linewidth}
            \centering            \includegraphics[width=\linewidth]{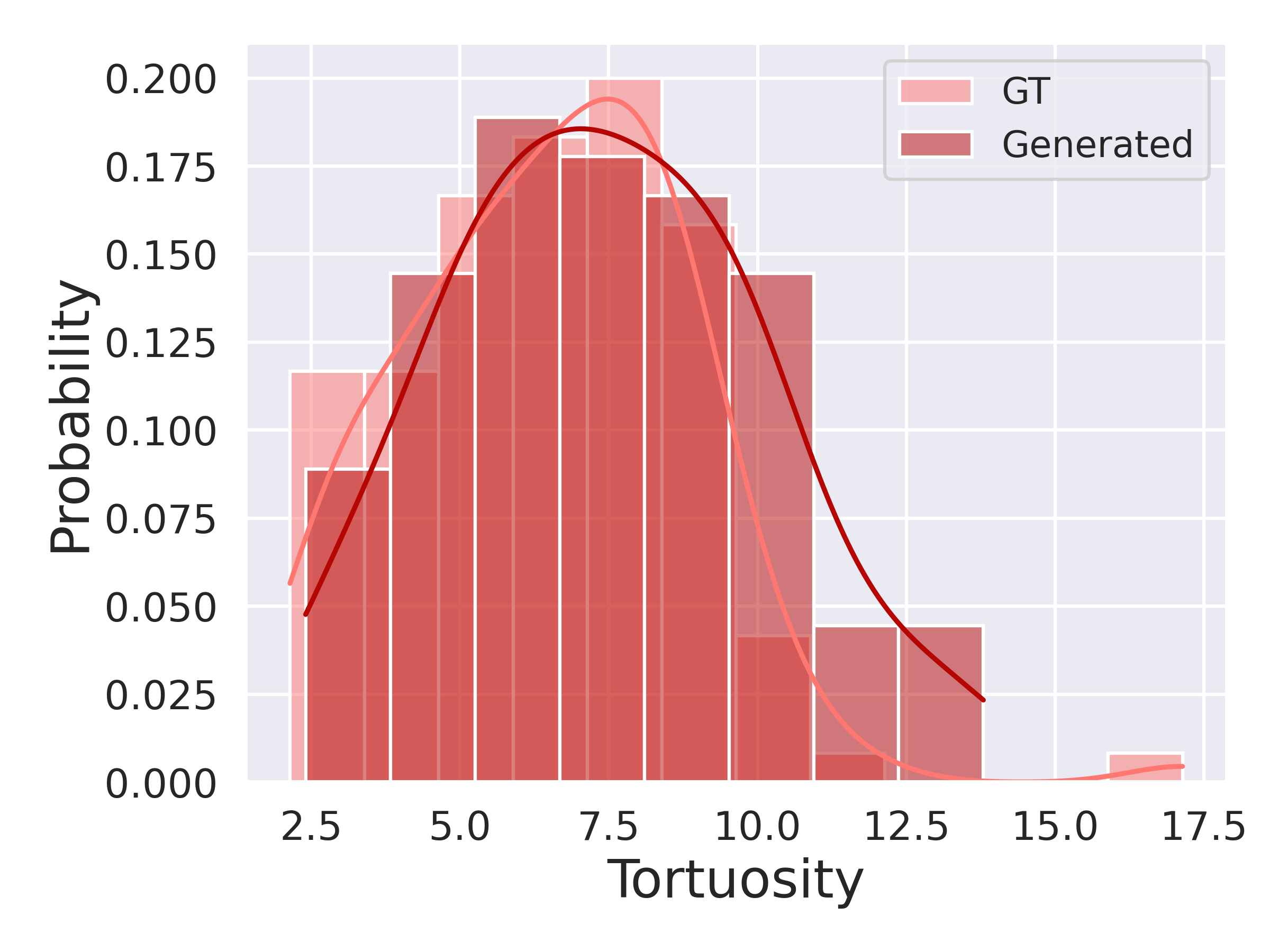}
            \caption{}
            \label{fig:tortuosity}
        \end{subfigure}
        \begin{subfigure}{0.49\linewidth}
            \centering            \includegraphics[width=\linewidth]{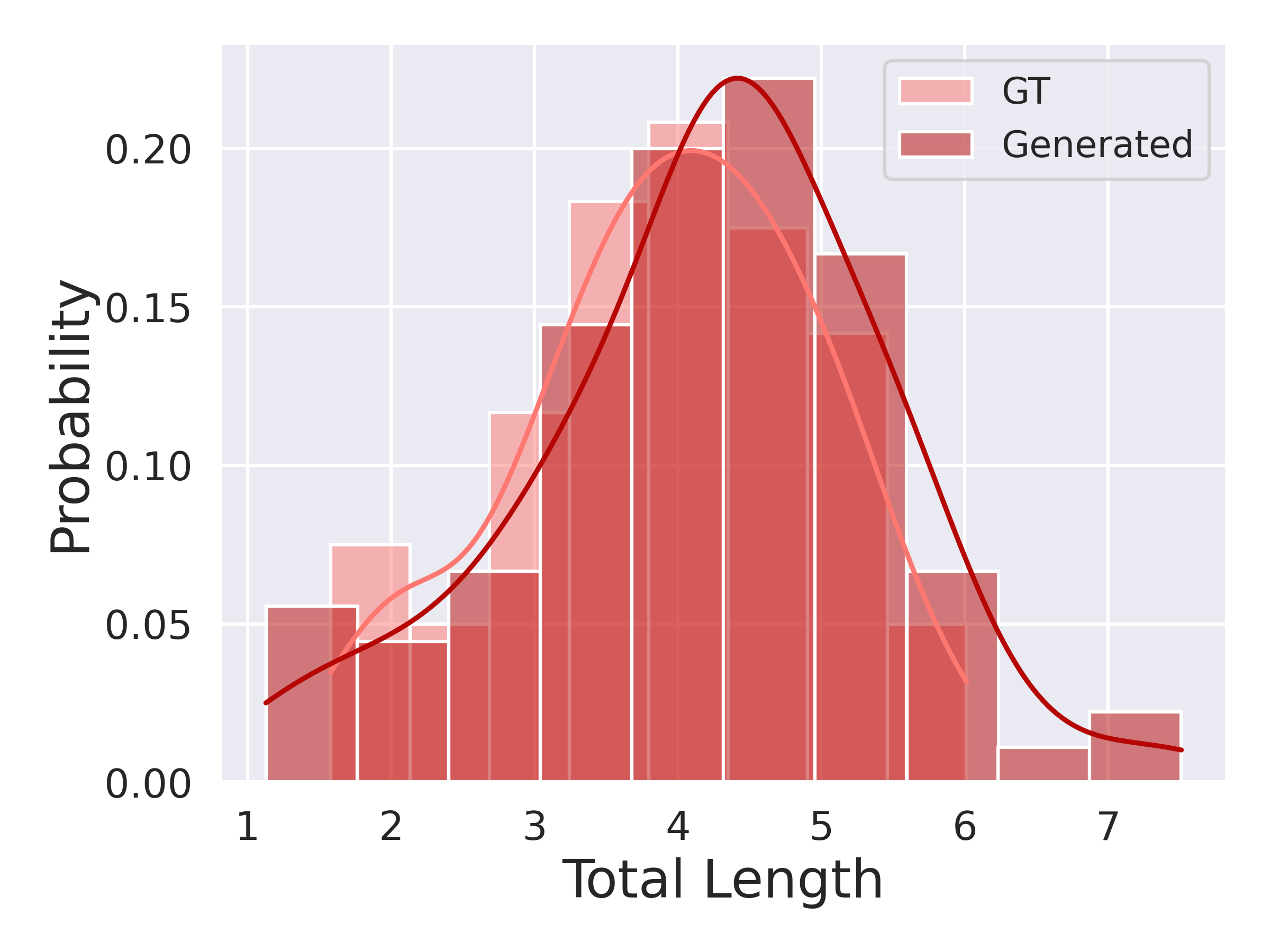}
            \caption{}
            \label{fig:tree_length}
        \end{subfigure}
    \label{fig:tree_distance}
    \caption{\textbf{Tree Statistics} computed on VascuSynth. (a) Distance between INRs of trees with similar bifurcations is low and increases as the complexity increases; (b) t-SNE plot of the space of trees as INRs; (c-d) Histograms of tortuosity, total length and average radius for ground truth and generated trees.}
\end{figure}

\begin{table}[h]
  \centering
  \setlength\tabcolsep{0pt}
  \begin{minipage}{.75\textwidth}
    \large
    \resizebox{\linewidth}{!}{
    \centering
        \begin{threeparttable}
          \caption{Quantitative results on tree structures present in different medical imaging modalities represented using INRs. We report the relative percentage error ($\%$) between the reconstructed signal and ground truth.
          }
            \begin{tabular}{lccc}
            
            \toprule
            {Modality} & {Rel. Error ($\%$) } & {Input Size} (MB) &{ INR Size} (MB)\\
             \midrule
             DRIVE (RGB)~\cite{staal2004drive}  & $ 0.018$ & $0.37_{\small \pm 0.0055}$ & $0.066_{\downarrow \times5.60}$ \\ 
             DRIVE (Mask)~\cite{staal2004drive} & $ 1.204$ & $0.02_{\small \pm 0.0013}$  &  $0.003_{\downarrow \times6.60}$ \\
             BraTS~\cite{menze2014multimodal} & $0.039$ & $68.11_{\small \pm 0.00}$ & $0.753_{\downarrow \times90.45}$ \\
             HAN-Seg~\cite{podobnik2023han} & $5.627$ & $12.1_{\small \pm 1.55}$ & $0.630_{\downarrow \times19.20}$ \\
            \bottomrule 
        \end{tabular}
        \label{tab:modality_inr}
        \end{threeparttable}
        }
  \end{minipage}%
  \begin{minipage}{.25\textwidth}
    \centering
    \large
    \resizebox{\linewidth}{!}{
    \begin{threeparttable}
        \caption{Quantitative results for novel tree generation on VascuSynth. $\uparrow$: higher is better; $\downarrow$: lower is better. For $1$-NNA, $50\%$ is ideal.}
            \begin{tabular}{lr}
                \toprule
                \rotatebox{0}{Metric} & Value \\
                \midrule
                MMD $\downarrow$ &  $13.36_{\small \pm 8.37}$ \\
                COV $\uparrow$ &  $0.46_{\small \pm 0.11}$ \\
                $1-$NNA $(\%)$ $\downarrow$ &  $87.49_{\small \pm 8.99}$ \\
                \bottomrule
                \end{tabular}
            \label{tab:novel_shape_transposed}
        \end{threeparttable}
    }
  \end{minipage}
\end{table}

\noindent \paragraph{\textbf{Space of INR-based Trees.}}
Once trees are represented via INRs, we study the space of INRs. 
The t-SNE plot in \Cref{fig:tsne_few} illustrates how trees with similar number of bifurcations, \ie topological complexities, tend to cluster in the 2D t-SNE space.
However, we observe cases where trees with the same number of branches are distant, which may be due to how t-SNE projects high-dimensional samples to 2D thus distort inter-sample distances.
Therefore, to further investigate, we compute the matrix of $L_2$ distances between INRs of VascuSynth trees with $i$ and $j$ bifurcations, where $i,j \in \{1,\dots,12\}$ (\cref{fig:l2_tree}). The reported distances are the average across $10$ trees per bifurcation count. We notice low ($\approx0$) values along the diagonal, indicating that trees with the same branch count have similar INRs. As the difference in branch count increases, the distance increases, visible by the color gradient in~\cref{fig:l2_tree}. 
Furthermore, for a constant difference in branch count (\eg, $|3-4|= 1 =|8-9|$), the dissimilarity is more pronounced with larger branch counts, despite the numerical difference being the same.

\begin{figure}
    \centering
    \begin{minipage}{\textwidth}
        \begin{subfigure}{\linewidth}
            \includegraphics[width=0.10\linewidth]{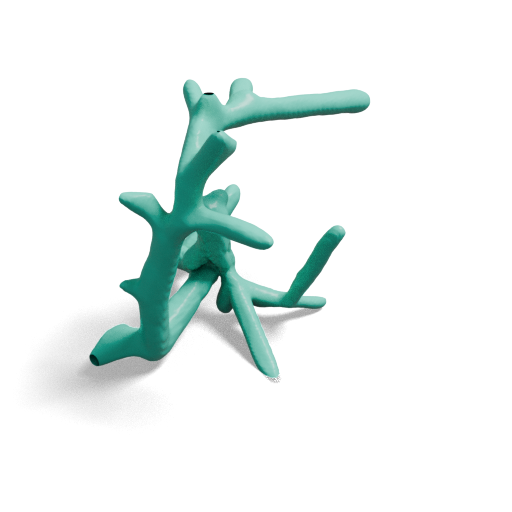}
            \includegraphics[width=0.10\linewidth]{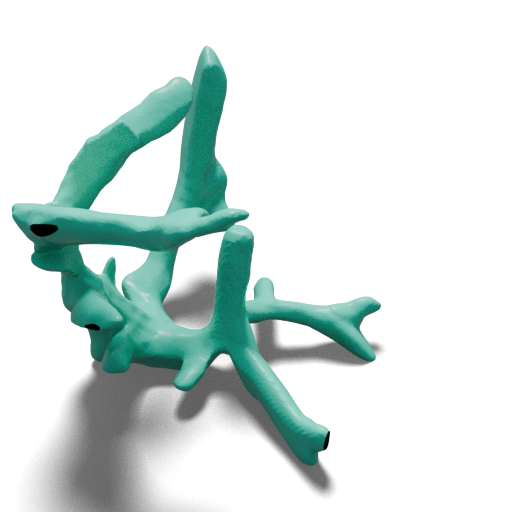}
            \includegraphics[width=0.10\linewidth]{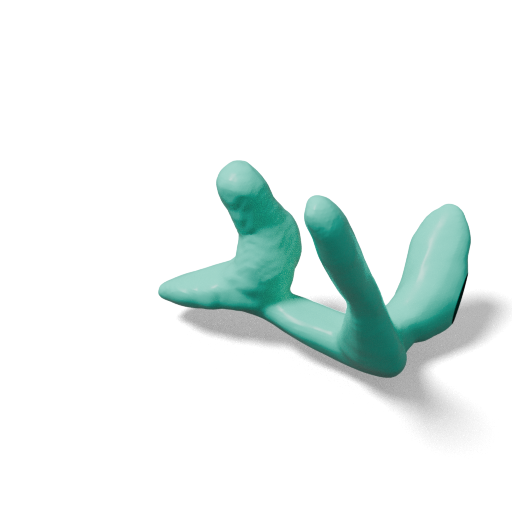}
            \includegraphics[width=0.10\linewidth]{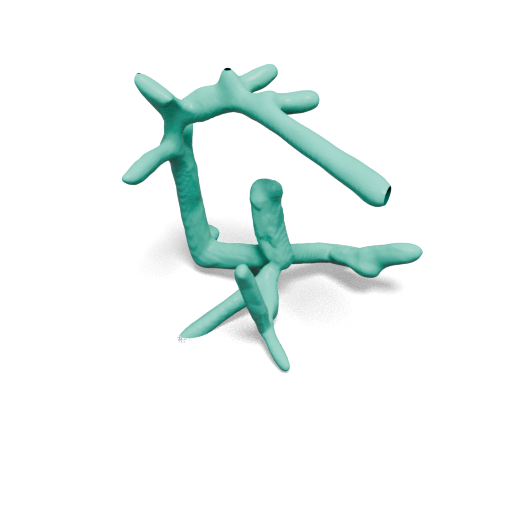}
            \includegraphics[width=0.10\linewidth]{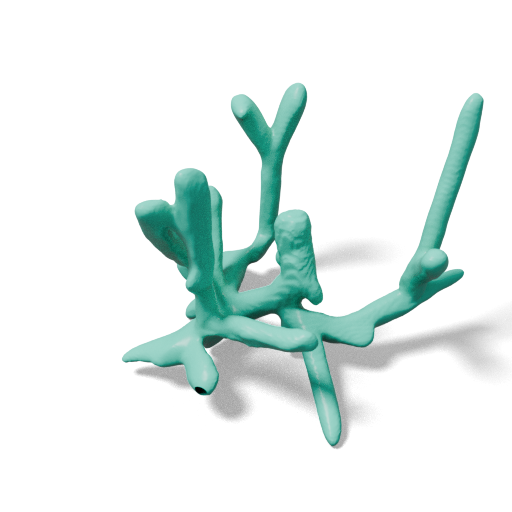} 
            \includegraphics[width=0.10\linewidth]{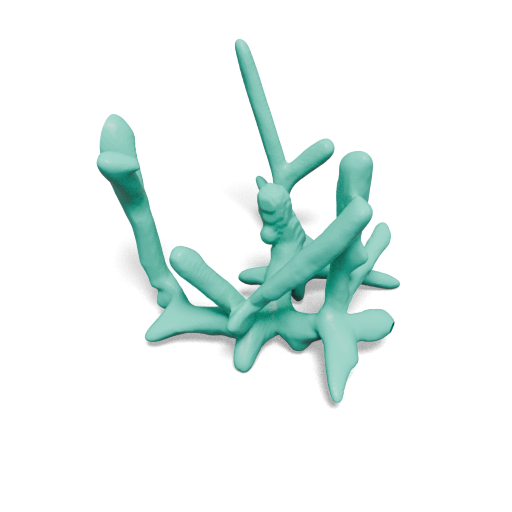} 
            \includegraphics[width=0.10\linewidth]{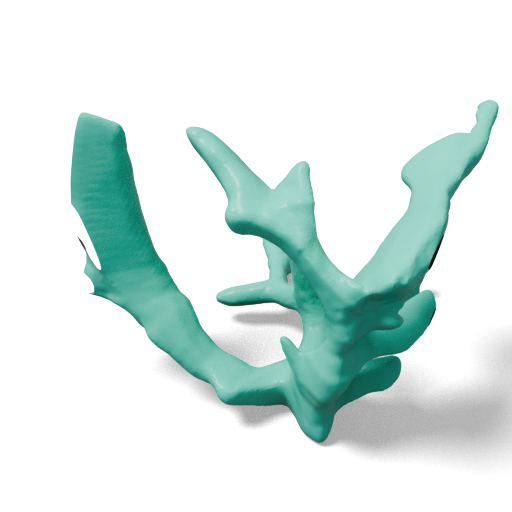}
            \includegraphics[width=0.10\linewidth]{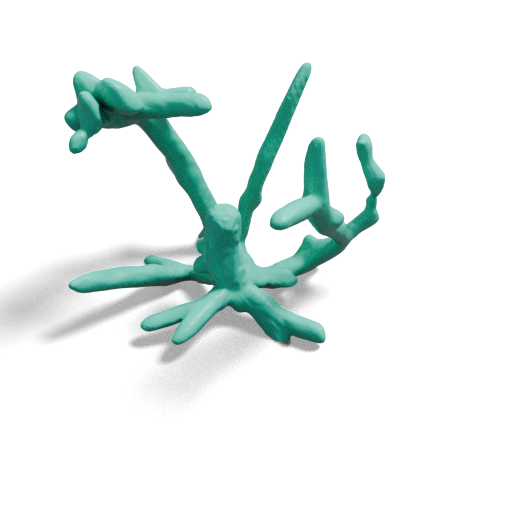}
            \includegraphics[width=0.10\linewidth]{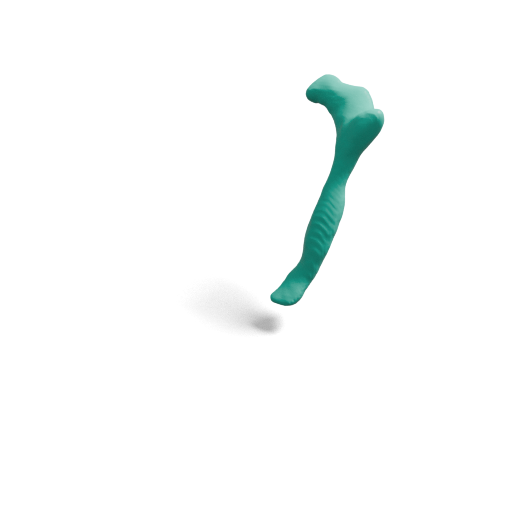} \\
            \includegraphics[width=0.10\linewidth]{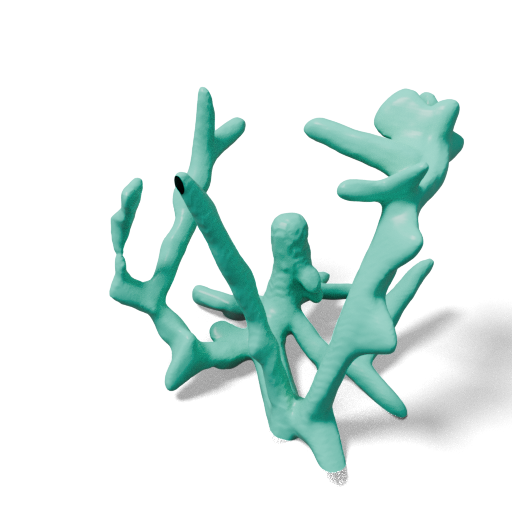} 
            \includegraphics[width=0.10\linewidth]{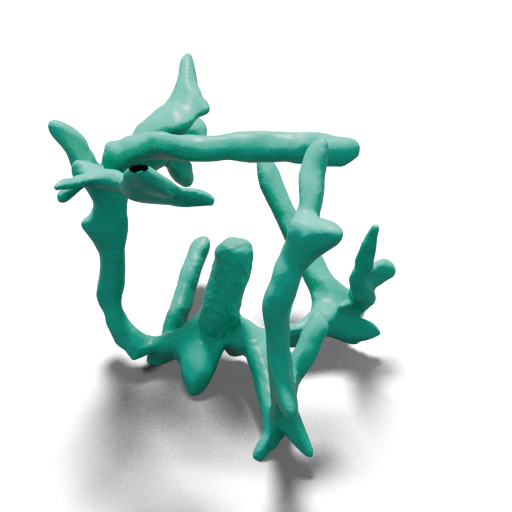}
            \includegraphics[width=0.10\linewidth]{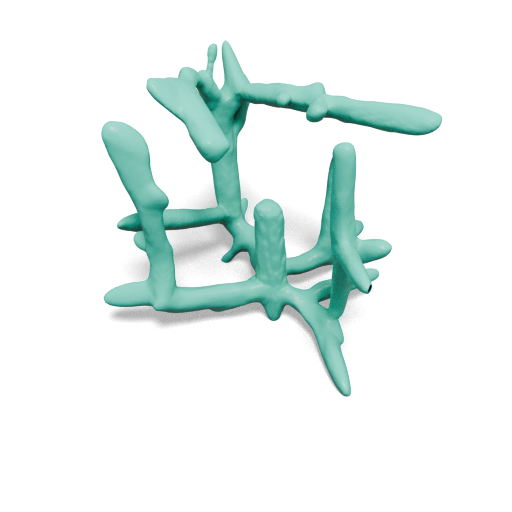}        
            \includegraphics[width=0.10\linewidth]{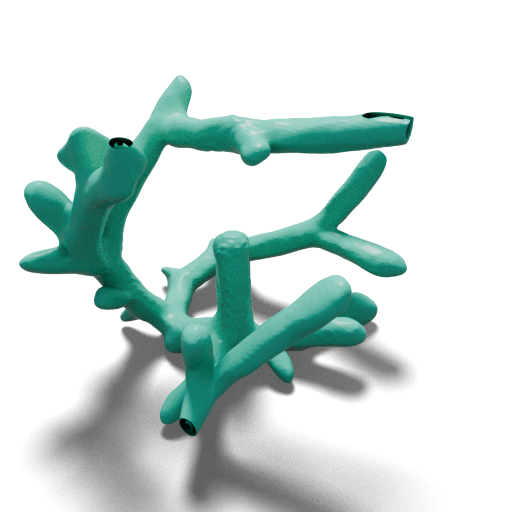}
            \includegraphics[width=0.10\linewidth]{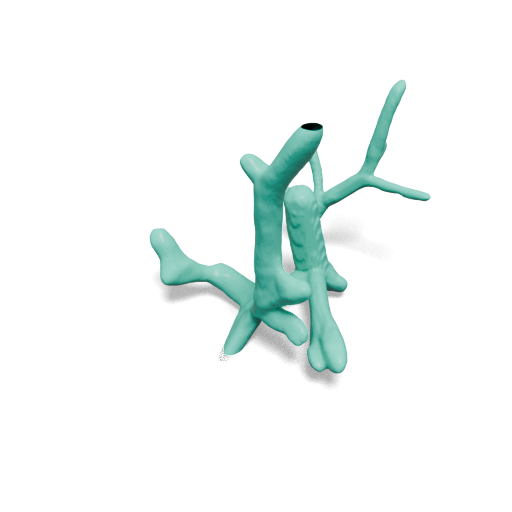}
            \includegraphics[width=0.10\linewidth]{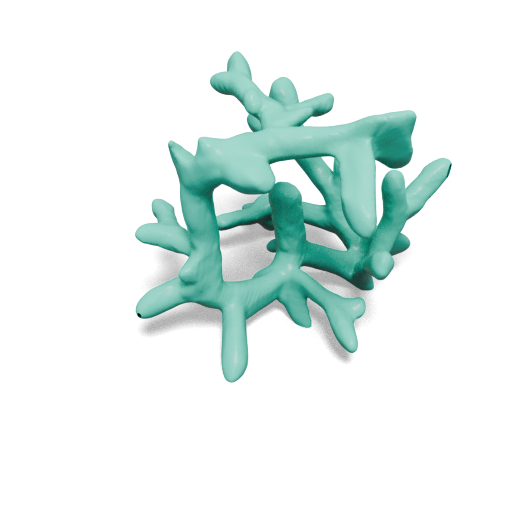}
            \includegraphics[width=0.10\linewidth]{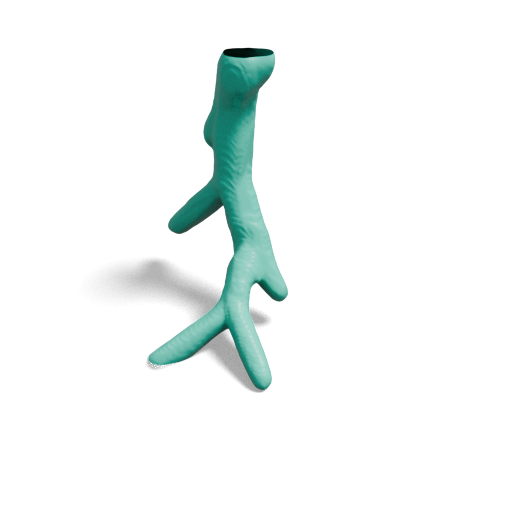}
            \includegraphics[width=0.10\linewidth]{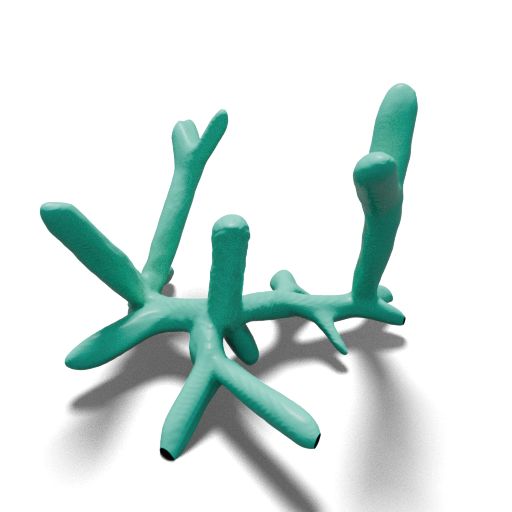}
            \includegraphics[width=0.10\linewidth]{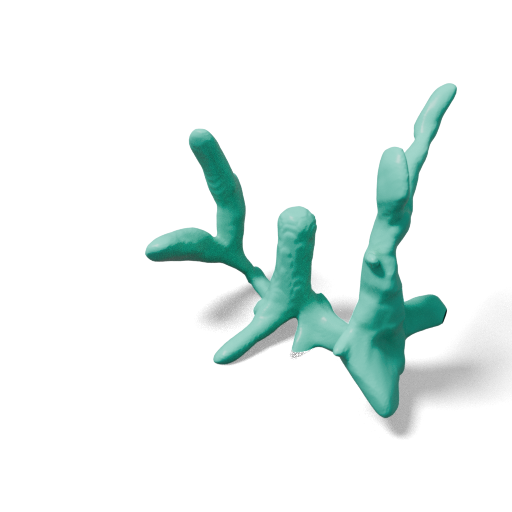}            
        \end{subfigure}
    \end{minipage}
    \caption{\textbf{Tree Synthesis.} Some novel INRs sampled from the diffusion model, visualized as meshes.}
    \label{fig:diffusion_gen}
\end{figure}

\noindent \paragraph{\textbf{Tree Synthesis.}}
Training a DDM on INR-based trees results in a model that captures the tree distribution and can be sampled using DDIM~\cite{song2020denoising} to generate novel trees. First, we show qualitative results of novel INR-based trees generated by our model in \Cref{fig:diffusion_gen}.
Next, since the evaluation of unconditional generation of tree-like structures can be challenging due to lack of direct correspondence to ground truth data, we follow~\cite{chou2023diffusion,erkocc2023hyperdiffusion} to quantitatively assess the samples generated using diffusion.
Specifically, we use minimum matching distance (MMD), coverage (COV), and 1-nearest neighbor accuracy (1-NNA) to measure quality, diversity, and plausibility, respectively.
Moreover, since the generations are random, we run the evaluation 3 times and report the mean values in~\cref{tab:novel_shape_transposed}.
Additionally, we follow~\cite{feldman2023vesselvae,wolterink2018blood} and report vessel-based metrics
in~\cref{fig:tortuosity,fig:tree_length}, where we see 
high similarities in
tortuosity and centerline length histograms between ground truth training data and synthesized trees.

\begin{figure}[h]
    \centering
    \raisebox{0.01\height}{\makebox[0.01\textwidth]{\rotatebox{90}{\makecell{\scalebox{0.6}{Circle of Willis}}}}}
    \includegraphics[width=0.9\textwidth,height=0.3\textwidth,keepaspectratio, trim={0 0.5cm 0 0.5cm}, clip]{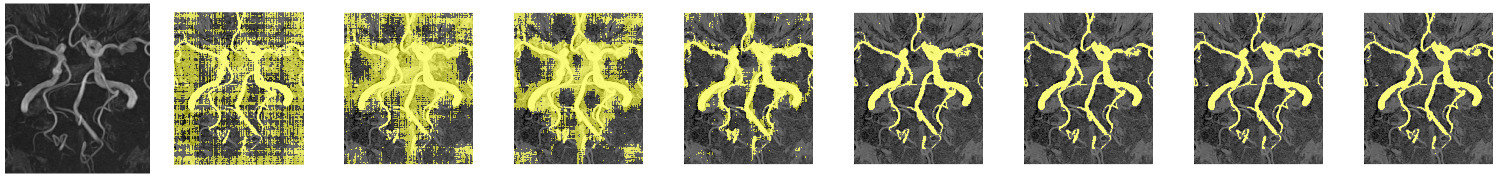} 
    \\
    \raisebox{0.01\height}{\makebox[0.01\textwidth]{\rotatebox{90}{\scalebox{0.6}{Whole Body MRA}}}}
    \includegraphics[width=0.9\textwidth, height=0.3\textwidth,keepaspectratio,trim={0 1.15cm 0 1.15cm}, clip]{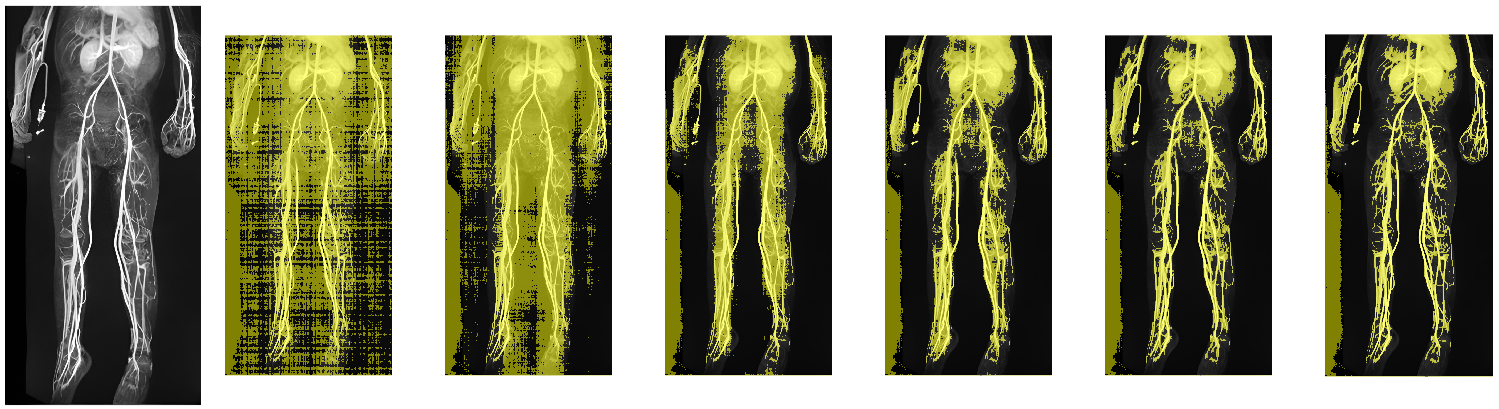} 
    
    \caption{\textbf{Segmentation}. The evolution of INR towards segmenting two vessel trees. Analogous to Mumford-Shah based segmentation~\cite{chan2001active}, we use a piecewise-constant version of the INR (via simple thresholding) as it is being optimized to faithfully represent the original image.}
    \label{fig:inr_seg}
\end{figure}

\noindent \paragraph{\textbf{INR-based Image Segmentation.}}
We present two proof-of-concept experiments on how INR representation is leveraged to perform vessel tree segmentation. \Cref{fig:inr_seg} depicts the evolution of INR-represented segmentation masks as they gradually fit to target vasculature in both CoW and WBMRA.

\noindent \paragraph{\textbf{Implementation Details.}}
We used PyTorch and ADAM optimizer with learning rate $\alpha=10^{-3}$ for all experiments.
To fit INRs to trees, we optimize the MLP architecture in~\cite{mescheder2019occupancy} for max $5k$ iterations on 12GB NVIDIA GeForce GTX TITAN X GPU, taking $\sim$1 minute/tree.
We train the diffusion model (128-D hidden size, 4 layers, and 4 self-attention modules, as in~\cite{peebles2022learning}) on 48GB NVIDIA A40 GPU for $6k$ epochs, 
with batch size 8, $0.9\times \alpha$ decay
per 200 epoch, $1k$ diffusion time-steps, %
linear noise scheduler $\in [10^{-4},10^{-2}]$, taking $\approx3$ days.

\section{Conclusion}
\label{sec:conclusion}

We presented \methodname, the first work to use implicit neural fields for faithful representation of topologically-complex anatomical trees and learnt their distribution via training a diffusion model in the space of neural fields. We demonstrated quantitatively and qualitatively the advantages of our method: versatility (\eg, 2D/3D; vascular/airway; simple/complex topology), lower memory footprint while achieving highly accurate reconstructions at arbitrary high resolutions; synthesis of plausible trees; and application to segmentation of medical images. Further, our representation is amenable to integration into deep learning pipelines and can be easily transformed into other shape representations. Future work may include integrating our method into more advanced deep segmentation pipelines while encoding semantic annotations of tree parts.

\section{Acknowledgements}
The authors are grateful to Paula Feldman for the provision of evaluation scripts used for computing tortuosity and other tree-specific metrics.
This research was enabled in part by support provided by the Natural Sciences and Engineering Research Council of Canada (NSERC), and the computational resources provided by Digital Research Alliance of Canada and NVIDIA Corporation.

\clearpage
\small
\bibliographystyle{splncs04}
\bibliography{miccai-short}

\end{document}

%% file: math_cmds.tex
\usepackage{amsmath,amsfonts,bm}

\def\eqref#1{equation~\ref{#1}}

\def\1{\bm{1}}

\def\vs{{\bm{s}}}

\DeclareMathAlphabet{\mathsfit}{\encodingdefault}{\sfdefault}{m}{sl}
\SetMathAlphabet{\mathsfit}{bold}{\encodingdefault}{\sfdefault}{bx}{n}

\DeclareMathOperator*{\argmin}{arg\,min}

%% file: commands.tex
\definecolor{green}{rgb}{0, 0.5, 0}
\definecolor{orange}{rgb}{0.8, 0.6, 0.2}
\definecolor{red}{rgb}{1.0, 0.0, 0.0}
\definecolor{teal}{rgb}{0.0, 0.4, 0.4}
\definecolor{purple}{rgb}{0.65,0,0.65}
\definecolor{saffron}{rgb}{0.95,0.75,0.2}
\definecolor{turquoise}{rgb}{0.0,0.5,0.5}
\definecolor{black}{rgb}{0.0, 0.0, 0.0}
\definecolor{gray}{rgb}{0.5, 0.5, 0.5}